# Stochastic Threshold Model Trees: A Tree-Based Ensemble Method for Dealing with Extrapolation

Kohei Numata[(1),§], Kenichi Tanaka[(1),§,*]

In the field of chemistry, there have been many attempts to predict the properties of unknown compounds from statistical models constructed using machine learning. In an area where many known compounds are present (the interpolation area), an accurate model can be constructed. In contrast, data in areas where there are no known compounds (the extrapolation area) are generally difficult to predict. However, in the development of new materials, it is desirable to search this extrapolation area and discover compounds with unprecedented physical properties. In this paper, we propose Stochastic Threshold Model Trees (STMT), an extrapolation method that reflects the trend of the data, while maintaining the accuracy of conventional interpolation methods. The behavior of STMT is confirmed through experiments using both artificial and real data. In the case of the real data, although there is no significant overall improvement in accuracy, there is one compound for which the prediction accuracy is notably improved, suggesting that STMT reflects the data trends in the extrapolation area. We believe that the proposed method will contribute to more efficient searches in situations such as new material development.

## 1 Introduction

One of the main purposes of regression analysis is to construct good statistical models of the form $\boldsymbol{y} = f(\boldsymbol{X})$. Such models are constructed from training data $\{(\boldsymbol{x}_1, y_1), (\boldsymbol{x}_2, y_2), \ldots, (\boldsymbol{x}_n, y_n)\}$ and used to predict the target value of a new sample, $(\boldsymbol{x}_{\text{new}}, y_{\text{new}})$. When $\boldsymbol{x}_{\text{new}}$ lies outside the range of the training data, the prediction is called an extrapolation. In searching for new materials in the field of chemistry, there are cases where it is desirable to search and make predictions in unknown areas, i.e., to extrapolate. However, extrapolation is generally more difficult than interpolation, for which $\boldsymbol{x}_{\text{new}}$ lies inside the range of the training data [1].

Many physicochemical properties involve monotonicity. For example, the spring length of the scale is proportional to its weight, polar molecules dissolve well in water, etc. Therefore, it is reasonable to try to extrapolate to physicochemical events in terms of monotonicity. Linear models can extrapolate to reflect the monotonicity of the training sample, but the prediction accuracy in interpolation is reduced in the case of the relationship between $x$ and $y$ is nonlinear. Nonlinear models have high prediction accuracy in interpolation, but the monotonicity observed from the training data is often not reflected to extrapolation.

To illustrate this, using simple 1-dimensional data, we have used Multiple Linear Regression (MLR) for linear models, Support Vector Regression (SVR) [2] and Random Forest (RF) [3] for nonlinear models and checked how the models make predictions, respectively. **Figure 1** shows the prediction results for linear data and **Figure 2** shows the results for nonlinear data. Blue dots denote the training data, and $y$-values are predicted in the range of $x$ on the graph. In the case of the linear data, MLR can extrapolate to reflect the trend of the training sample. SVR fits the training sample in interpolation, but in extrapolation, the predictions converge to the mean of the training sample as it moves away from the training sample. In the case of the linear data, the extrapolation is also inappropriate, depending on the parameter settings, as shown in **Figure 1**. For RF, the interpolation fits the training sample, but the predictions between training samples are step-like. In extrapolation, the prediction is the value of the nearest neighboring sample group. **Figure 1** confirms that the RF does not reflect the trend of the training samples either. With the nonlinear data, no valid extrapolation can be obtained using MLR, SVR, or RF (**Figure 2**).

As mentioned above, the conventional nonlinear regression methods cannot extrapolate adequately in certain cases. These examples illustrate the problem due to the property of the models, and are expected to improve the extrapolation behavior. In particular, when making predictions for new materials, it is desirable to use a nonlinear regression method that can predict the sample values in a region where there are few known data. This study aims to develop a nonlinear regression method that can make reasonable extrapolations while maintaining the accuracy of the conventional

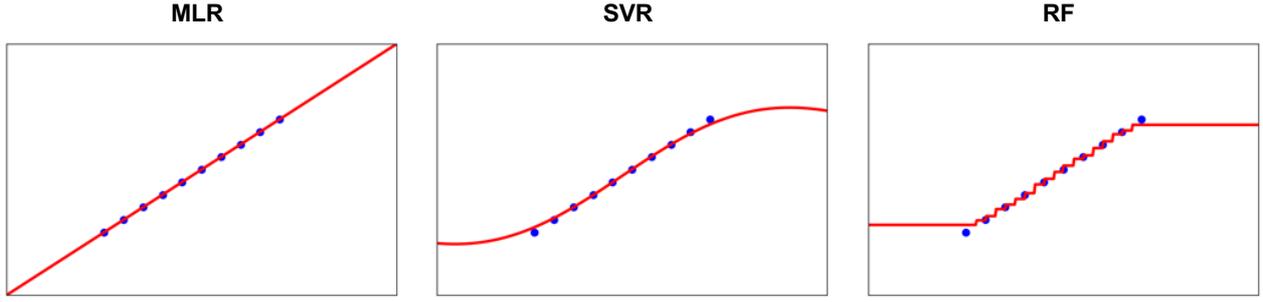

Figure 1. Predictions using linear data.

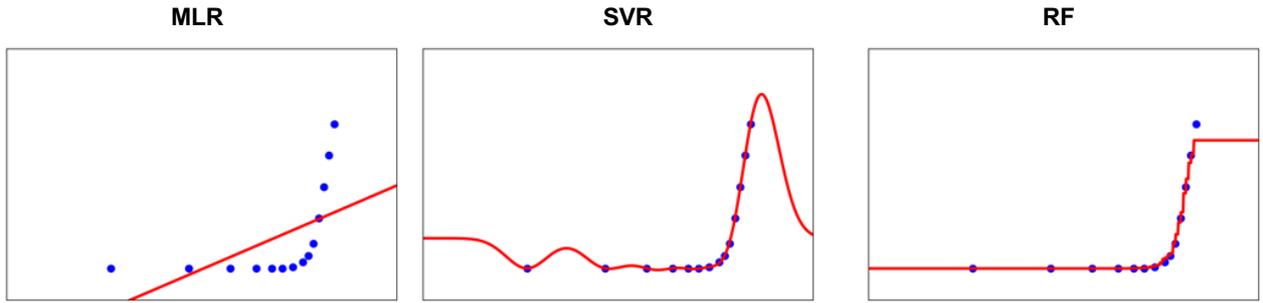

Figure 2. Predictions using nonlinear data.

interpolation method and without adjusting the parameters.

To achieve this goal, we proposed Stochastic Threshold Model Trees, STMT. We change two parts of the RF algorithm, specifically the tree partitioning method and the model construction method at each terminal node after partitioning. For the latter, a method has been proposed for constructing a linear regression model from sample groups of terminal nodes in a regression tree (Model Trees [4]), and this approach is adopted in the present study.

The remainder of this paper is organized as follows. We introduce the RF algorithm and the details of STMT in Sec. 2. The results of experiments using STMT with artificial data are presented in Sec. 3, and those with a real dataset are given in Sec. 4. Finally, we conclude this paper in Sec. 5.

## 2 Conventional methods

### 2.1 Support Vector Regression (SVR)

SVR is a smooth nonlinear regression model. The regression model is often constructed based on the following equation:

$$f(x) = \sum_{n} \alpha_n \exp(-\gamma \|x - x_n\|^2) + \beta,$$

where $\gamma$ is constant, and $\alpha_n$ and $\beta$ are determined by learning.

When constructing the regression model, SVR considers an allowable error, $\varepsilon$. Samples that deviate more than $\varepsilon$ from the regression curve, such as the red points in **Figure 3**, are called support vectors. SVR constructs its regression model by superimposing a Gaussian distribution centered on each support vector.

In the case of data for which the density depends on the data section, the appropriate hyper-parameters differ in each section, and a good model cannot be constructed. In addition, the extrapolated value cannot always reflect the trend of the training data, because the prediction converges to the bias value, $\beta$.

As shown in **Figure 1**, extrapolation cannot give appropriate predictions if improper parameter adjustment is performed, and prediction and extrapolation become inappropriate for data with mixed densities, as shown in **Figure 2**.

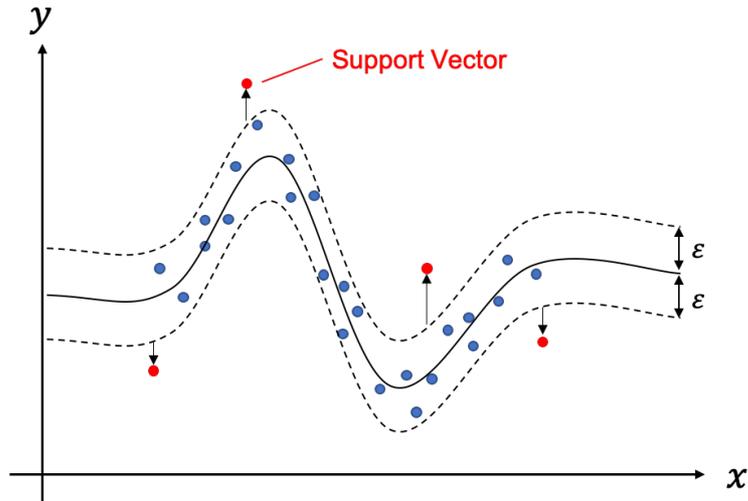

**Figure 3. Example of support vector in SVR model.** Samples with errors greater than $\varepsilon$ are called support vectors, and the regression model is constructed using these samples.

**2.2 Random Forest (RF)**

RF is an ensemble method using multiple regression trees that have low predictive performance by themselves. In RF, bootstrap sampling [5] is first applied to the training data. Bootstrap sampling is a method in which, when there are $n$ samples of the training data, sampling with replacement is performed $n$ times to create new sub-datasets. A regression tree is then constructed for each sub-dataset. In each regression tree, the data are recursively divided at thresholds where the variance of the target of the training data becomes small. Moreover, the average value of each partition area is output as a predicted value (**Figure 4**). Partitioning is performed based on the feature that gives the optimal threshold in the dataset. The midpoints between samples are selected as threshold candidates. In RF, random feature selection is performed in addition to bootstrap sampling. The average of the predicted values obtained from the regression trees is the output value of the RF. Regression trees are easy to overfit individually, but RF offers improved generalization by integrating various regression trees.

RF constructs a model by aggregating a number of regression trees, and returns the average value of samples in the local section as the prediction result. Therefore, the extrapolated value becomes close to the value of the nearest training samples. Additionally, when the data density is low, the thresholds may be concentrated at the midpoint between samples, even when multiple regression trees are integrated. This leads to the construction of a step-like regression model because the midpoint is used as the threshold when dividing the data. These properties of RF are illustrated in the examples shown in **Figure 1** and **Figure 2**.

**2.3 Model Trees (MTs)**

Model trees (MTs) combine linear regression models and regression trees to create a hybrid model that produces better predictions. Thus, MTs are analogous to piecewise linear regression.

**Figure 4** shows how MTs are constructed. Basically, partitioning is performed at the point where the error of linear models after partitioning is minimized. Finally, an output value is obtained by connecting the prediction results of the linear models at the terminal nodes.

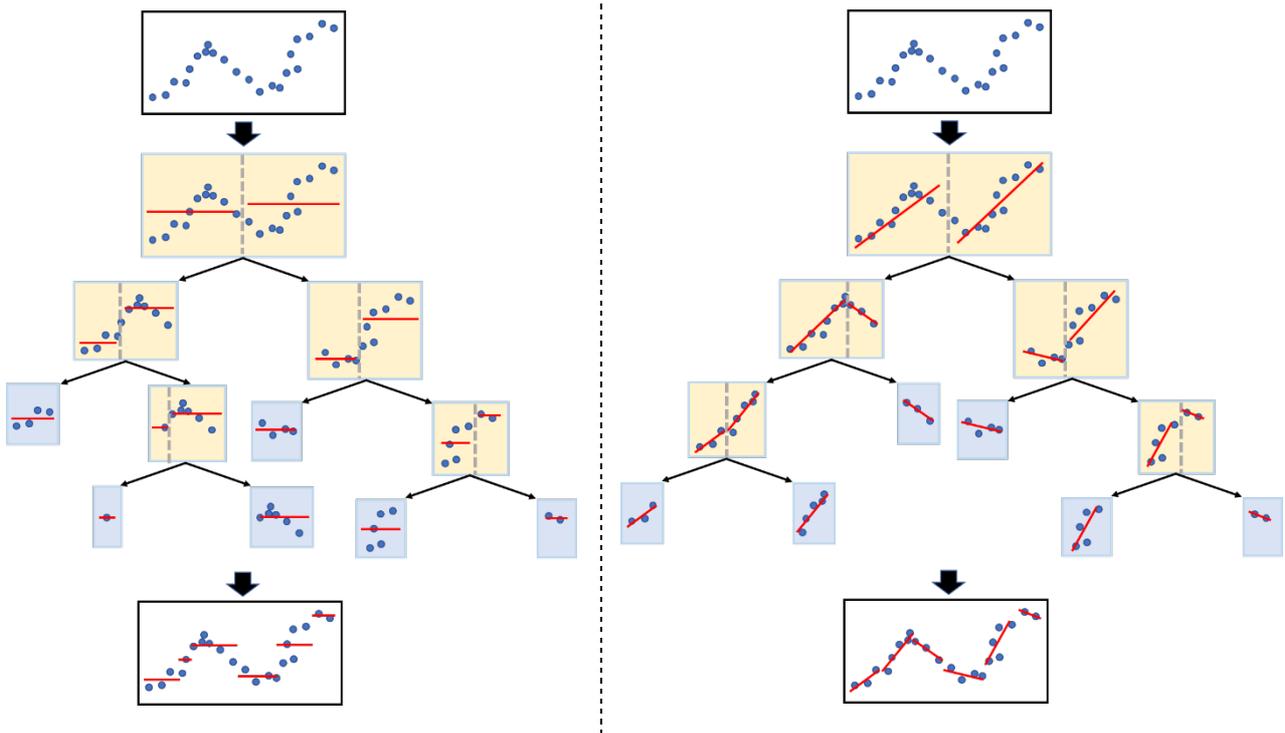

**Figure 4. Difference between a regression tree and a model tree.** The figure on the left illustrates the model construction of a regression tree, and right shows of a model tree.

## 3 Proposed method: Stochastic Threshold Model Trees (STMT)

### 3.1 Overview of STMT

In STMT, the following two aspects of RF are modified to obtain a reasonable extrapolation result.
1. Determine thresholds probabilistically, rather than fixing them at the midpoint between samples.
2. Apply linear regression to the sample group of terminal nodes and use the prediction result as the prediction value of each regression tree, rather than using the average value of terminal nodes. In other words, replace regression trees with model trees.

The first modification introduces some randomness to the selection of the thresholds, and makes it possible to obtain a smooth prediction curve, even for regions with sparse data. The second modification makes it possible to obtain extrapolated values that reflect the trend in high-data-density regions using linear regression.

### 3.2 Details of STMT

3.2.1 Determination of thresholds

In the regression tree used in STMT, the thresholds are determined using a normal distribution. Focusing on two samples that are adjacent to each other for a certain feature, let us compare how the threshold is determined in RF and in STMT. **Figure 5** shows the difference in the determination of the threshold between the two methods.

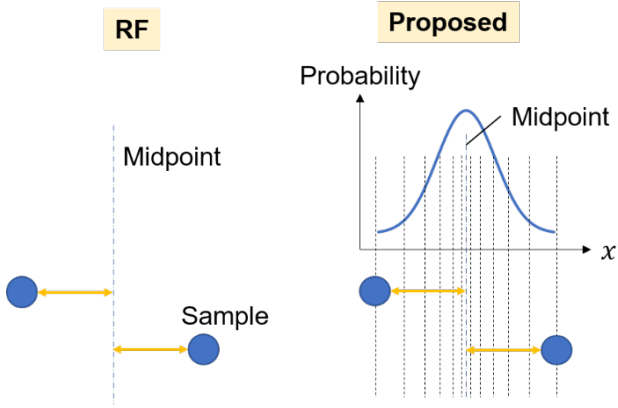 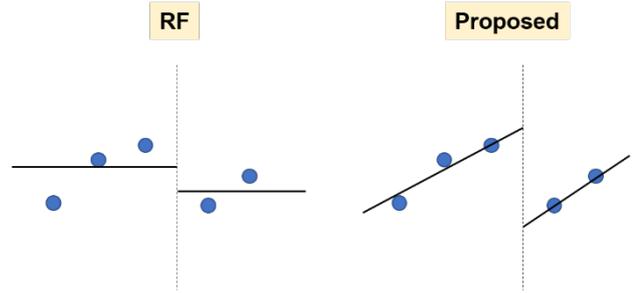

**Figure 5. Difference between RF and the proposed method in determination of the thresholds.** In STMT, we consider a normal distribution centered on the midpoint of two samples, and determine the threshold according to this probability.

**Figure 6. Difference between RF and the proposed method in construction of a model at each partition area.** In STMT, linear regression is applied as well as MTs.

In RF, the midpoint between two samples is selected as a candidate for the threshold (as shown on the left of **Figure 5**). In contrast, STMT selects the threshold so as to have a probability according to a normal distribution centered on the middle point between samples (as shown on the right of **Figure 5**). Assuming that the feature of interest in the two samples has values of $x_1$ and $x_2$, respectively, this normal distribution is expressed by the following equation:

$$N\left(\frac{x_1 + x_2}{2}, \left(\frac{x_2 - x_1}{2k}\right)^2\right)$$

where $k$ is the hyper-parameter.

As described above, smooth prediction results can be obtained while maintaining the properties of RF to some extent, as the selection is more likely to be closer to the middle point while assigning a degree of randomness to the selection. The relationship between the distance from the midpoint to the sample and the standard deviation, which determines the shape of the normal distribution, is controlled by the hyper-parameter.

3.2.2 Application of linear regression method

In STMT, linear regression is applied to each partition area. **Figure 6** shows the difference between the construction of a model at the terminal node in RF and in STMT. In RF, the average value of the sample group at the terminal node is the predicted value, whereas in STMT, the linear regression result is used as the predicted value.

When applying linear regression, MLR is used because no parameter adjustment is necessary. The following two methods can be considered when introducing the linear regression.
1. For consistency with RF, after determining the thresholds that minimize the sum of squared errors with respect to the mean value, linear regression is applied to sample groups of terminal nodes.
2. Linear regression is applied to each disjoint region for all threshold candidates, and the threshold that minimizes the sum of squared errors between the measurement value and the prediction value is selected (as with MTs).

In the first method, the number of linear regression models created is the same as the number of nodes. In the second method, it is necessary to create as many regression models as the number of samples for each threshold, which incurs a high computational cost. Therefore, we use the first method in this study.

When the number of features is large, the number of samples required to construct the linear models may be smaller than the number of features. This leads to the generation of an unstable model. Therefore, when building a linear model at the end of a regression tree in STMT, you can choose to use only the variables selected in the process of partitioning or to use all the variables.

The existence of outliers when applying the linear models may lead to the construction of inappropriate models, causing the predictions to deviate from the true values. For this reason, when aggregating the regression trees, STMT basically uses the median value as the output rather than the mean value.

3.2.3 Hyper-parameters

The hyper-parameters of STMT are based on the parameters of RF implemented in scikit-learn [6], and the same names are used for similar parameters. The modified parameters and new parameters are as follows.

The first parameter is the minimum number of samples required at a leaf node. In scikit-learn, the minimum number of samples or the ratio of the number of samples to that of training samples is specified. In STMT, the minimum number of samples or the ratio of the number of samples to that of features used in the training is specified. Determining the minimum number of samples based on the number of features prevents the number of samples at the node from being less than the number of features, thereby preventing the linear model from becoming unstable.

The second parameter is the value of $k$, which determines the shape of the normal distribution (see Sec. 3.2.1). We set $k\sigma$ to be equal to the distance between the midpoint and the sample, where $\sigma$ is the standard deviation of the normal distribution.

The final parameters concern the choice of performing a specific process when applying the linear models. As described at the end of Sec. 3.2.2, the linear models can become unstable or inappropriate. Whether or not to select features for constructing linear models or to use the median value when aggregating regression trees in order to deal with these problems are determined by hyper-parameters. These hyper-parameters are particularly effective when the data have a large number of features or outliers.

## 4 Experiments – Artificial data

In this section, we verify whether STMT behaves as expected when applied to artificial data. We prepared data with one variable and two variables that can easily be visualized. The minimum number of samples required to be at a node was set to two in the one-variable data and three in the two-variable data.

**4.1 Results – One-variable artificial data**

Using a data series in which one area contains no samples, i.e., discontinuous data, we successively set $k$ to 1, 3, 5, and 7. In addition, RF and MT were applied for comparison. The results are shown in **Figure 7**, where the blue dots indicate the training data and the red line is the predicted output.

Although RF was able to predict the interpolation points correctly, it could not obtain extrapolation results reflecting the trend of interpolation. In contrast, MT and STMT gave extrapolated predictions that reflected the trend of the samples near the extrapolation area.

With RF and MT, step-like prediction results appeared in the low-density area. With STMT, the step-like prediction was smoothed out by changing how the thresholds were determined, as described in Sec. 3.2.1. We compared the prediction results given by various values of $k$. When the distance between the midpoint and the sample was $1\sigma$, the prediction for the central two samples was observed to deviate because the threshold was relatively likely to be selected from outside of the two target samples. Thus, it is necessary to specify an integer of three or more as the value of $k$ to obtain appropriate thresholds. In subsequent experiments, we set $k = 5$.

A comparison of the predictions given by MLR, SVR, RF, MTs, and STMT for various one-variable artificial datasets is shown in **Figure 8**. From this figure, it is visually apparent that STMT gives predictions that are smooth and reflect the trends of the data.

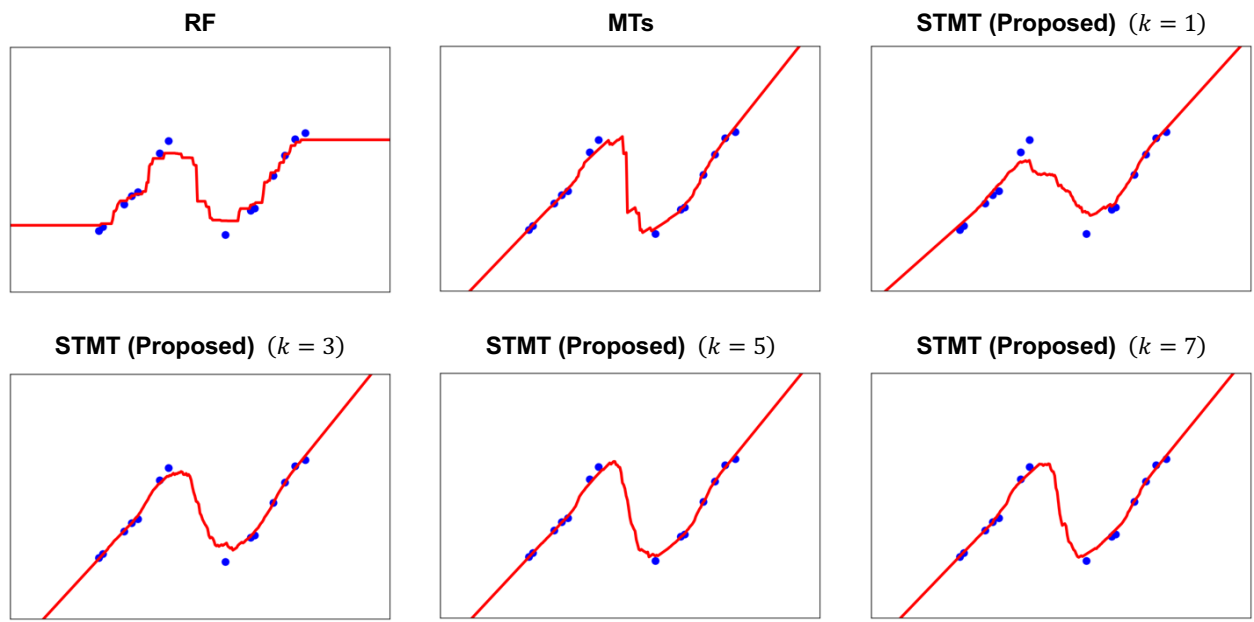

**Figure 7. Predictions using each method with discontinuous, one-variable artificial data.** In addition to RF and MTs, the prediction results given by STMT with various values of $k$ are shown.

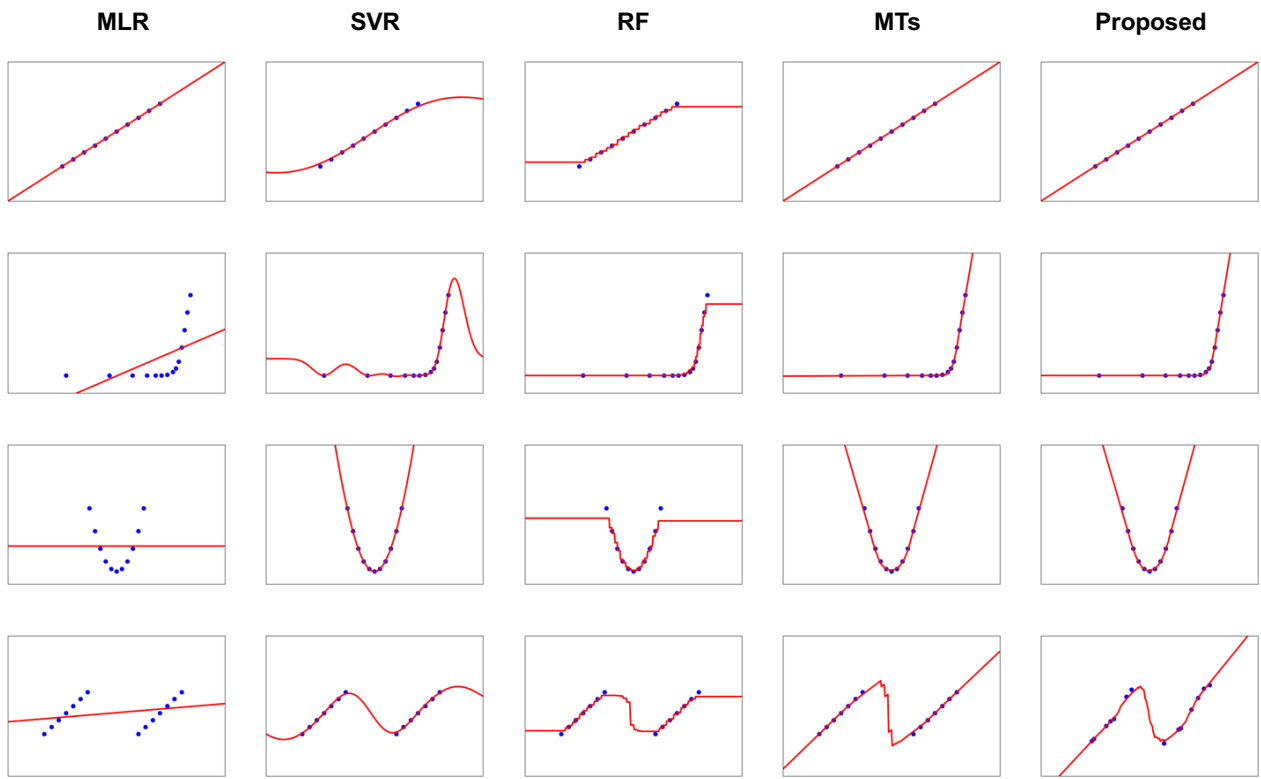

**Figure 8. Prediction results given by MLR, SVR, RF, MTs, and the proposed method for various datasets.**

**4.2 Results – Bivariate artificial data**

Next, we examined the difference in behavior between the conventional methods and STMT using bivariate artificial data, as illustrated in **Figure 9**. These data are based on $y = x_1^2 + x_2^2$, with some added noise. The training data and the test data are plotted as blue and yellow points, respectively, and the predictions given by each method were compared with the yellow points.

The results are shown in **Figure 10**. As in the one-variable case, interpolation is done well by SVR and RF, but problems appear in the case of extrapolation. With STMT, by reflecting the trend of the data, it was possible to obtain reasonable predictions for both interpolation and extrapolation.

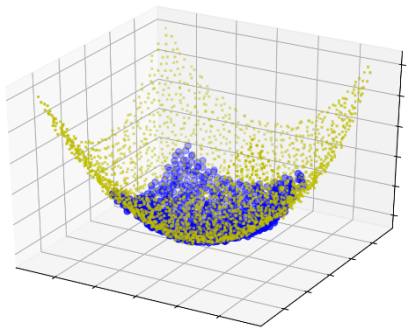

**Figure 9. Bivariate artificial data.**

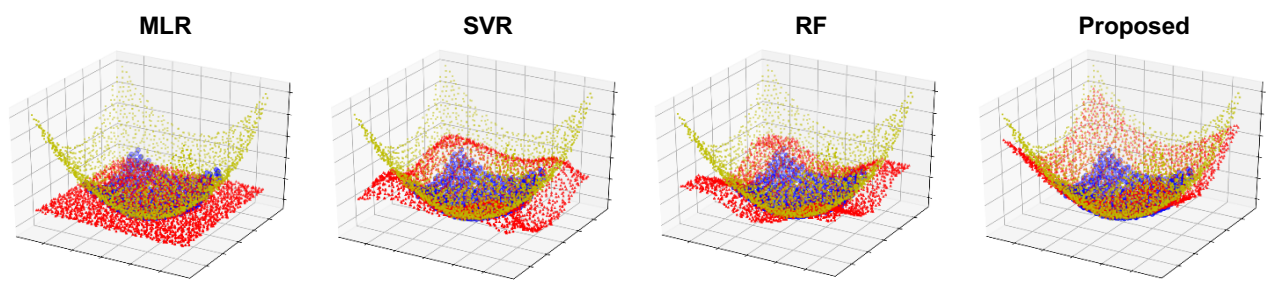

**Figure 10. Predictions using bivariate artificial data.** The predictions given by MLR, RF, SVR, and the proposed method are presented.

## 5 Experiments – Real data

This section reports the experimental results from using STMT with real data. For comparison, the real data were also applied with four conventional methods, MLR, RF, SVR, and Extra Trees (ETs) [7], which is a tree-based ensemble method.

### 5.1 Datasets

Water solubility (log S) data for 1,290 compounds [8] were used to verify the performance of STMT. For each compound, 196 structural descriptors were calculated using RDKit [9]. The following preprocessing steps were then performed.
- Remove descriptors with regard to octanol–water partition coefficient (log P).
- Remove descriptors with constant values.
- Remove one of the highly correlated descriptor pairs.

There is a strong correlation between log S and log P. Thus, log P can significantly improve the prediction accuracy, making it difficult to compare accuracy between methods. Therefore, log P was removed.

To confirm the prediction accuracy of the extrapolations made by STMT, the data for the extrapolation area were created through the following procedure (**Figure 11**).
1. Divide the dataset randomly into training data and test data at a ratio of $7:3$.
2. Apply the One Class Support Vector Machine (OCSVM) [10], an anomaly detection method, to the training data and obtain the separation hyperplane between the normal and abnormal values.
3. Divide the test data into those in the interpolation/extrapolation areas using the separation hyperplane determined in step 2.
4. Construct a regression model from the training data.
5. Predict the log S value of the test data using the model constructed in step 4, and confirm the prediction accuracy of interpolation and extrapolation.

The random division into training data and test data was performed 10 times by changing the random seed value, and the average value of 10 trials was taken as the final value for comparison.

OCSVM can calculate the signed distance from the separating hyperplane. A larger distance indicates a higher probability of a normal value, whereas a smaller distance suggests a greater likelihood of an abnormal value (outlier). Hereinafter, this signed distance is referred to as the "decision function." Normally, a sample with a negative decision function is considered to be an outlier.

Based on the decision function, the test data were divided into those in the interpolation/extrapolation areas and outliers. **Figure 12** shows an example of the division of the test data based on the decision function. In the experiments, the decision function generally took a value in the range $[-10, 10]$, and the test data were divided according to the following criteria.
- Interpolation area:    $0 \leq d < 10$,
- Extrapolation area:    $-8 \leq d < 0$,
- Outliers:              $-10 \leq d < -8$,

where $d$ denotes the value of the decision function.

**Table 1** lists the number of data in each category for all 10 divisions. In each case, the number of data in the interpolation area was around 300, that in the extrapolation area was ~60–80, and the number of outliers was ~10. Outliers were excluded, and the prediction accuracy for the data in the interpolation/extrapolation area was confirmed.

In this analysis, the minimum number of samples required for a leaf node was set to be equal to the number of features used for partitioning in the regression tree. The use of these features in the construction of linear regression models at the terminal nodes makes it possible to construct good models.

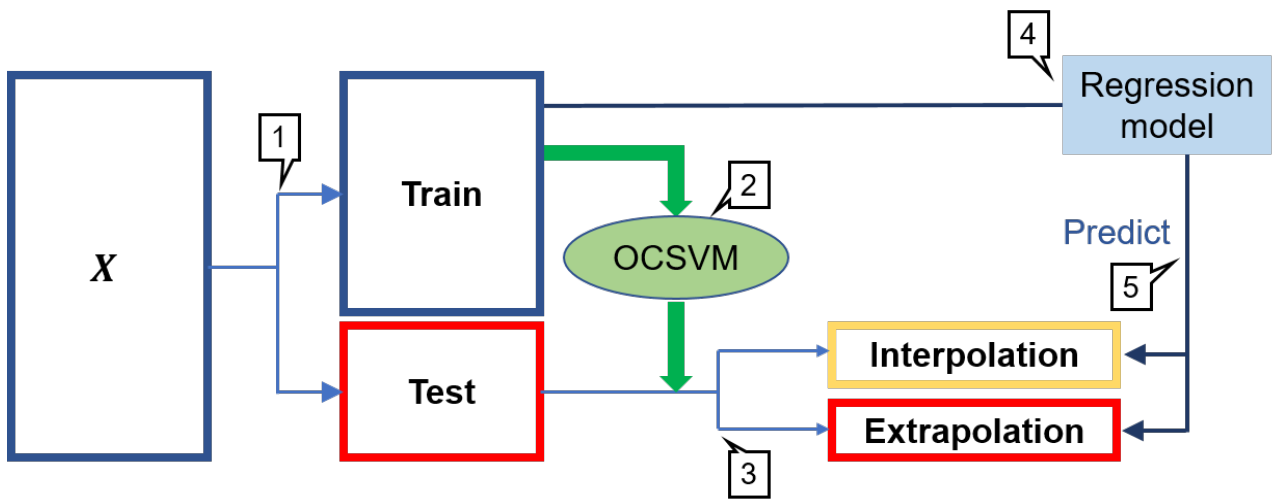

**Figure 11. Data division based on the features using OCSVM.** OCSVM was applied to the training data, and the test data were divided into interpolation/extrapolation areas based on the results.

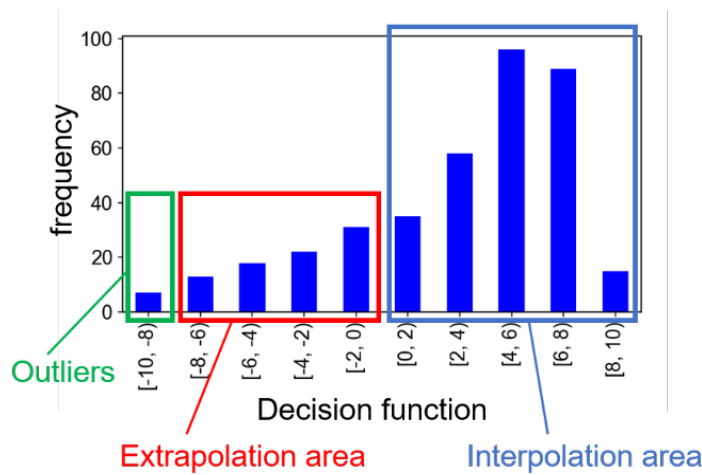

**Figure 12. Data division based on decision function.** $0$ and $-8$ were set as the boundaries between the interpolation and extrapolation areas and between the extrapolation area and the outliers, respectively.

**Table 1. Number of compounds in the data in the interpolation/extrapolation areas, and number of outliers.**

| Random seed value | Interpolation area | Extrapolation area | Outliers |
|---|---|---|---|
| 0 | 296 | 84 | 7 |
| 1 | 310 | 68 | 9 |
| 2 | 294 | 79 | 14 |
| 3 | 295 | 78 | 14 |
| 4 | 293 | 85 | 9 |
| 5 | 296 | 82 | 9 |
| 6 | 299 | 77 | 11 |
| 7 | 305 | 78 | 4 |
| 8 | 305 | 76 | 6 |
| 9 | 308 | 65 | 14 |

**Table 2. Prediction results of log S data analysis.**

| Method | Interpolation | | Extrapolation | |
|---|---|---|---|---|
| | $R^2$ | MAE | $R^2$ | MAE |
| MLR | $0.9118 \pm 0.0058$ | $0.4531 \pm 0.0129$ | $0.7999 \pm 0.0429$ | $0.7469 \pm 0.0485$ |
| RF | $0.9106 \pm 0.0103$ | $0.4329 \pm 0.0212$ | $0.8371 \pm 0.0352$ | $0.6573 \pm 0.0526$ |
| SVR | $0.9315 \pm 0.0079$ | $0.3814 \pm 0.0192$ | $0.8317 \pm 0.0401$ | $0.6534 \pm 0.0588$ |
| ET | $0.9212 \pm 0.0069$ | $0.4047 \pm 0.0150$ | $0.8510 \pm 0.0344$ | $0.6298 \pm 0.0439$ |
| Proposed | $0.9101 \pm 0.0089$ | $0.4248 \pm 0.0182$ | $0.8433 \pm 0.0354$ | $0.6313 \pm 0.0406$ |

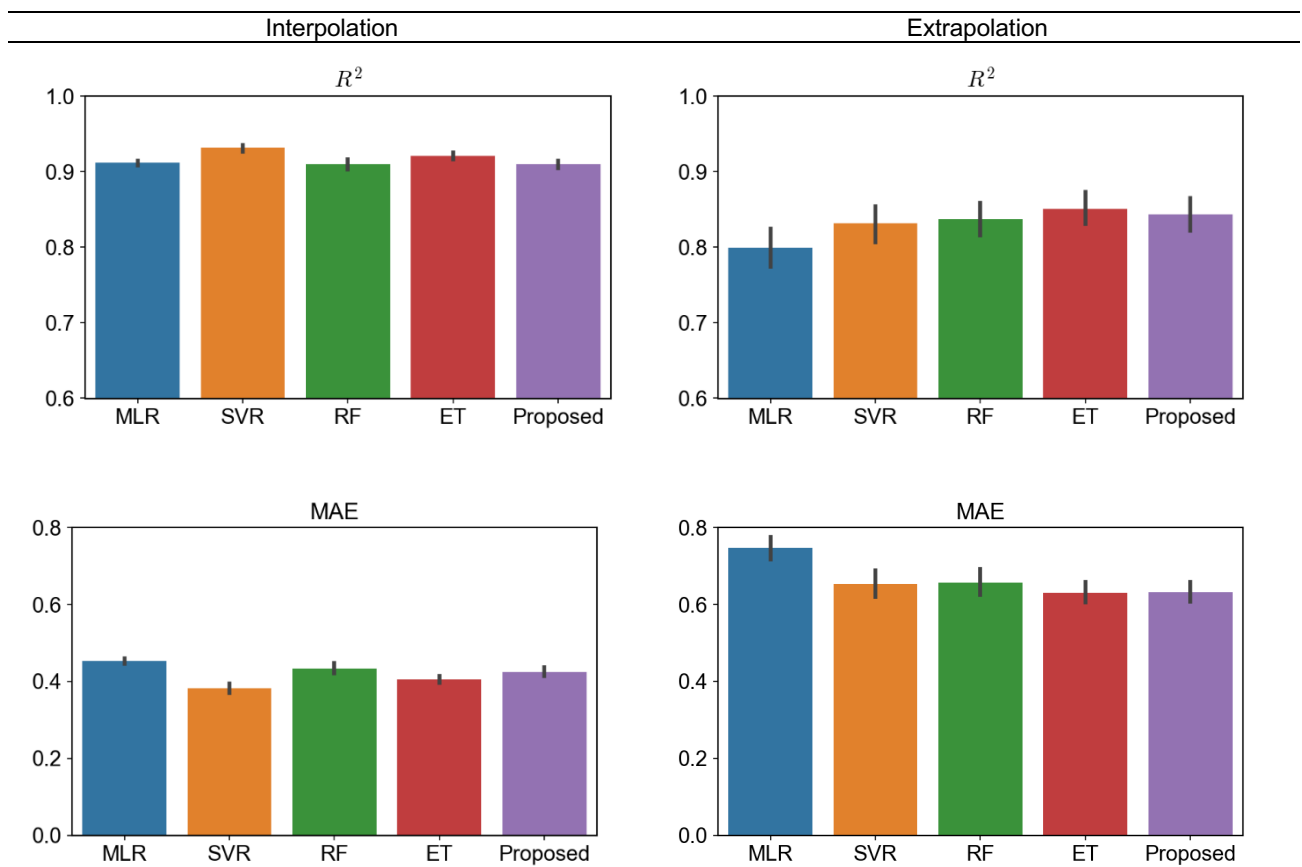

**Figure 13. Prediction results (bar plot) of log S data analysis.** The left-hand figures show the prediction accuracy of interpolation, and the right-hand figures show the prediction accuracy of extrapolation.

## 5.2 Results

The prediction results are presented in **Table 2** and **Figure 13**. As evaluation indices, the coefficient of determination ($R^2$) and the mean absolute error (MAE) were used. The error bar in each graph in **Figure 13** indicates the 95% confidence interval according to a $t$-test.

Regarding interpolation, the performance of STMT was inferior to that of SVR and ETs. The extrapolation performance of the proposed approach was the second-most accurate after ETs in terms of the average precision, but there was no significant difference when the standard deviation was added.

There were some differences between the methods in terms of the predictions for each compound. **Figure 14** plots the data in the extrapolation area, with the calculated values on the horizontal axis (True) and the predicted values of each method on the vertical axis (Pred). It can be seen that STMT achieved improved accuracy for the compound with the lowest log S value (denoted by the red circle in the figures) compared with the other methods, especially RF, SVR, and ETs. This compound is a kind of polychlorinated biphenyl (PCB), the structure of which is shown in **Figure 15**.

PCBs are manmade, mainly oily chemicals. They have the characteristics of being relatively insoluble in water, having a high boiling point, and being difficult to decompose under heating. For the 40 PCBs in the training data, the number of chlorine atoms and log S values are plotted in **Figure 16**. In this figure, the PCBs included in the data in the extrapolation area (for example, **Figure 15**) are indicated by a red dot. From this, it can be seen that the compound in **Figure 15** has a log S value outside the range of the training data, and that the log S value tends to decrease as the number of chlorine atoms increases.

It is considered that STMT accurately predicted the compounds shown in **Figure 15** by reflecting the tendency for a monotonic decrease in the log S value as the number of chlorine atoms increased in the extrapolation. In contrast, RF, SVR, and ETs encountered situations where appropriate extrapolation could not be performed, such as when the predicted value was within the range of the log S values of the training data.

This suggests that STMT is effective when the tendency of the data in the extrapolation area is easy to understand, such as when it involves monotonicity.

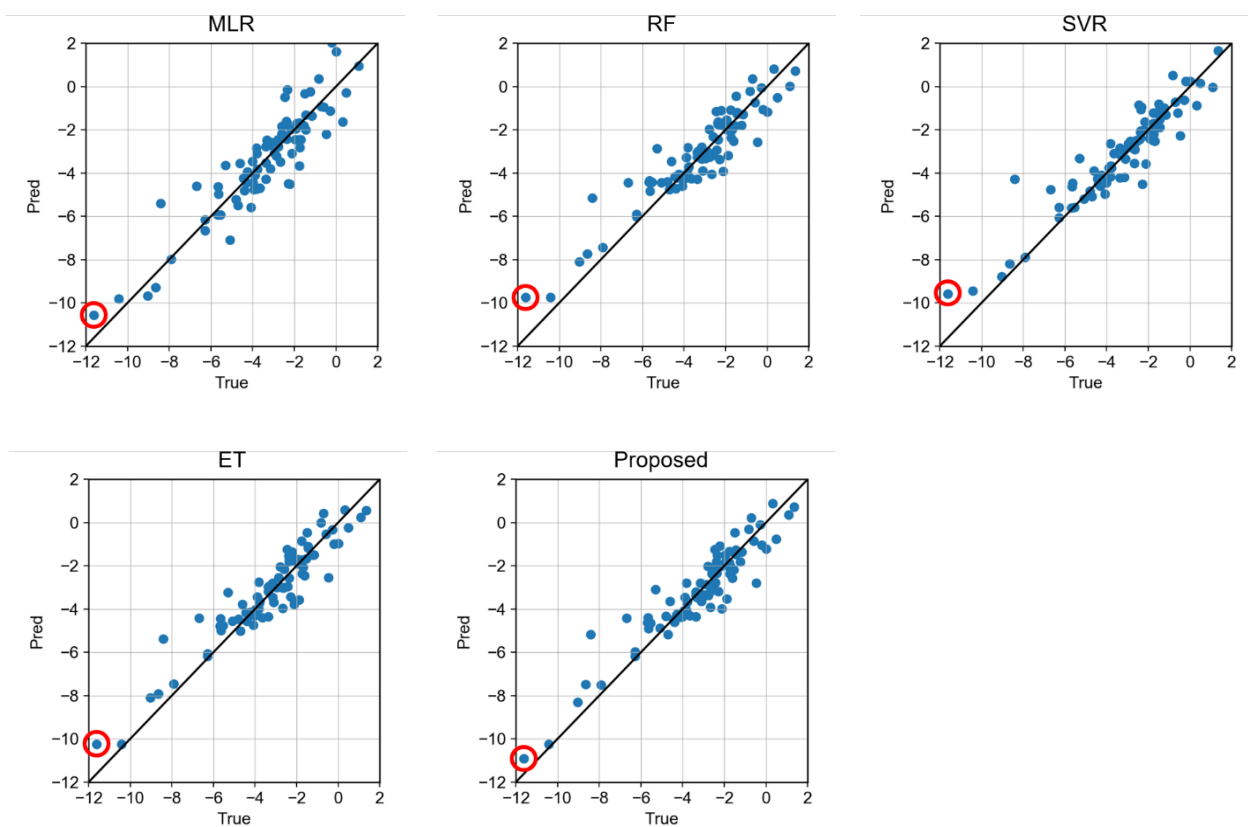

**Figure 14. Plots of true values and predicted values for each method.** Predictions that are closer to the diagonal are more accurate.

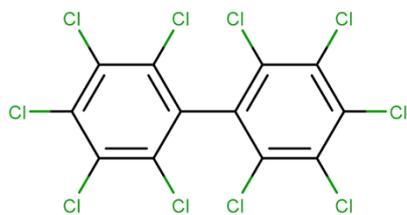 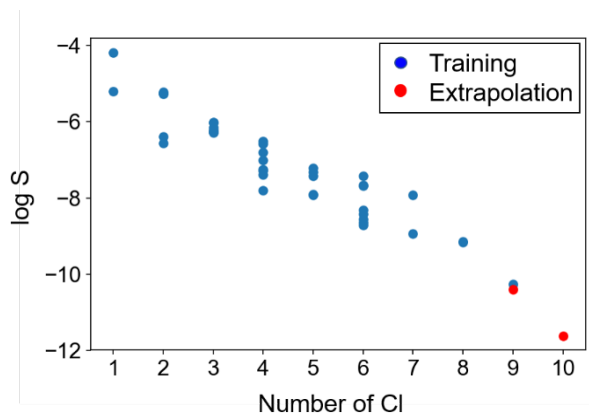

**Figure 15. Compound with the lowest log S value.** This is a kind of polychlorinated biphenyl (PCB).

**Figure 16. log S vs number of Cl.** The log S value tends to decrease as the number of chlorine atoms increases.

## 6 Conclusion

In this study, we have described STMT, a method for performing proper extrapolation. In STMT, the RF algorithm was modified and its extrapolation behavior was changed. The changes covered two points, namely how to determine the threshold when constructing the regression tree and how to construct the model at the terminal node.

To verify the accuracy of STMT, we conducted case studies using artificial data and real data. In the first case study, the behavior of STMT was confirmed using one- and two-variable artificial data. Hyper-parameters suitable for determining the threshold were determined from the results using the one-variable data. Analysis of the two-variable data confirmed that STMT behaved as expected.

In the second case study, we verified the accuracy of STMT using water solubility data. In this case, OCSVM was used to divide the data into interpolation/extrapolation areas, and the accuracy was confirmed for each area. The results showed that STMT was not significantly better at extrapolation than conventional methods. However, a certain compound exhibited improved accuracy compared with the other nonlinear regression methods. This compound had the same partial structure as other compounds in the training data, and STMT could reflect these trends in the extrapolation. This suggests that STMT can perform extrapolation more appropriately than conventional methods when the trend of the data is easy to understand, such as in the case of monotonicity. We believe that this method can be applied to the design of experiments in order to enable more efficient searches in situations such as the development of new materials.

## Author Contributions

K.N. wrote the manuscript, and all authors helped finalize the text. K.T. first designed the basic concept of the proposed method, STMT. K.N. mainly developed the code to implement STMT, and K.T. assisted. K.N. verified the effectiveness of STMT, and K.T. provided guidance on implementation and verification.

## Acknowledgments

We wish to acknowledge Prof. Funatsu, Professor of Chemical System Engineering, The University of Tokyo, for his help in providing a research environment. We thank Stuart Jenkinson, PhD, from Edanz Group (https://en-author-services.edanzgroup.com/) for editing a draft of this manuscript. We use following free software; python, anaconda, jupyter notebook, Visual Studio Code, numpy, scikit-learn, joblib, pandas, matplotlib, seaborn, we thanks for all contributors of these projects.